\pdfoutput=1

\documentclass[11pt]{article}

\usepackage[preprint]{acl}

\usepackage{times}
\usepackage{latexsym}

\usepackage[T1]{fontenc}

\usepackage[utf8]{inputenc}

\usepackage{microtype}

\usepackage{inconsolata}

\usepackage{graphicx}

\usepackage{csquotes}
\usepackage{booktabs}
\usepackage{url}
\usepackage{multirow}

\title{TartuNLP at SemEval-2025 Task 5: Subject Tagging as Two-Stage Information Retrieval}

\author{Aleksei Dorkin \and Kairit Sirts \\
        Institute of Computer Science \\
        University of Tartu \\
        \texttt{\{aleksei.dorkin, kairit.sirts\}@ut.ee} \\ }

\begin{document}
\maketitle
\begin{abstract}
We present our submission to the Task 5 of SemEval-2025 that aims to aid librarians in assigning subject tags to the library records by producing a list of likely relevant tags for a given document. We frame the task as an information retrieval problem, where the document content is used to retrieve subject tags from a large subject taxonomy. We leverage two types of encoder models to build a two-stage information retrieval system---a bi-encoder for coarse-grained candidate extraction at the first stage, and a cross-encoder for fine-grained re-ranking at the second stage. This approach proved effective, demonstrating significant improvements in recall compared to single-stage methods and showing competitive results according to qualitative evaluation.
\end{abstract}

\section{Introduction}

SemEval-2025 Task 5 aims to produce a technical solution to annotate a large collection of documents with relevant subject tags~\cite{dsouza-EtAl:2025:SemEval2025}. Subject tags come from the GND (Gemeinsame Normdatei in German or Integrated Authority File in English) subject taxonomy. For example, an article dealing with seismic resistance of industrial buildings may have subject tags such as ``Industrial Plant Technology (General)'' and ``Earthquake Safety Construction Engineering''.

Systematic and precise metadata annotation, subject tagging in particular, is essential for digital collections of documents to be usable and useful for the end-users as information retrieval and knowledge discovery sources. However, producing such metadata, especially at scale, requires an immense amount of specialist time and effort.
For instance, a recent prototype at the National Library of Estonia, Kratt, found that even with machine assistance catalogers still needed to validate many tags, highlighting the labor bottleneck~\cite{asula2021kratt}.

The core idea behind our system is that given a tag definition and a document to tag, it should be possible to predict whether the given tag is suitable for the document by measuring how similar their representations are using a pre-trained language model. This can be thought of as an information retrieval problem. 

Bi-encoder and cross-encoder models are two common ways to approach this problem~\cite{reimers-gurevych-2019-sentence, Nogueira2019PassageRW, karpukhin-etal-2020-dense}, both with their strengths and weaknesses. A bi-encoder processes documents and queries independently to produce their vector representations. Then, cosine similarity is measured between the document and query vectors to score relevance. Bi-encoder allows for efficient retrieval in large document collections as the document representations can be pre-computed. The downside, however, is that it is impossible to capture token-level similarities between documents and queries with this approach. Meanwhile, a cross-encoder processes document and query pairs simultaneously, capturing such similarities and producing more fine-grained similarity scores. The computational cost of using a cross-encoder is significantly larger compared to that of a bi-encoder, making it generally unsuitable for large document collections. Accordingly, we combine both in a single two-stage system to produce the optimal result.

\begin{table*}[h]
    \centering
    \begin{tabular}{lcc}
        \toprule
        Team Name & Average Recall (tib-core-subjects) & Average Recall (all-subjects) \\
        \midrule
        \textbf{RUC Team} & \textbf{0.6568} & 0.5856 \\
        \textbf{Annif} & 0.5899 & \textbf{0.6295} \\
        LA2I2F & 0.5794 & 0.4821 \\
        DUTIR831 & 0.5599 & 0.6045 \\
        icip & 0.4976 & 0.5302 \\
        Team\_silp\_nlp & 0.4939 & 0.1271 \\
        \textit{TartuNLP (ours)} & 0.4049 & 0.3818 \\
        JH & 0.2252 & 0.1677 \\
        last\_minute & 0.2099 & - \\
        Homa & 0.2030 & - \\
        TSOTSALAB & 0.0667 & - \\
        DNB-AI-Project & - & 0.5631 \\
        jim & - & 0.4686 \\
        NBF & - & 0.3224 \\
        \bottomrule
    \end{tabular}
    \caption{Average recall scores for tib-core-subjects and all-subjects subtasks of quantitatitative evaluation. Sorted by tib-core-subjects score, best scores and teams on each subtask are highlighted with bold font.}
    \label{tab:quantitative}
\end{table*}

Our system scored closer to the middle of the quantitative leaderboard (Table~\ref{tab:quantitative}); however, it surpassed many of the participants with a similar rank in the qualitative leaderboard (Table~\ref{tab:recall_teams}), especially in Case 1, where ``technically correct, but irrelevant'' tags were counted in favor of the teams. This is likely due to our system not considering the tag-to-tag relations, thus missing a more complex knowledge structure in the dataset and being unable to distinguish more intricate cases. More specifically, we consider each document/tag pair independently of each other. However, we hypothesize that some subject tags may be mutually exclusive.

We release the code for training and inference and the models.\footnote{\url{https://github.com/slowwavesleep/llms4subjects-submission}}

\section{Background}

In the shared task, the participants were given a collection of English and German documents containing a title and an abstract. The goal was to recommend the top N subjects most relevant from a predefined set of subjects. Each subject was represented by an alphanumeric code, a name in German, and an optional definition in German. Training and validation data provide ground truth tags for each document.

Two subtasks were offered to the participants: a complete collection of subjects (tibkat) and a smaller subset of the collection (tibkat-core). The former also contained a larger number of documents.
In our experiments, we used the smaller tibkat-core dataset. However, we submitted the final results for both subtasks.

\begin{table*}[h]
    \centering
    \begin{tabular}{l l c c}
        \toprule
        Rank & Team Name & Average Recall Case 1 & Average Recall Case 2 \\
        \midrule
        1  & DNB-AI-Project  & 0.7175 & 0.6406 \\
        2  & RUC Team        & 0.7155 & 0.6254 \\
        3  & DUTIR831        & 0.6966 & 0.6125 \\
        4  & Annif           & 0.6797 & 0.5866 \\
        5  & LA212F          & 0.6659 & 0.5718 \\
        6  & \textit{TartuNLP (ours)}        & 0.6649 & 0.5291 \\
        7  & jim            & 0.6601 & 0.5620 \\
        8  & icip           & 0.6310 & 0.5241 \\
        9  & NBF            & 0.6117 & 0.4622 \\
        10 & last\_minute    & 0.3971 & 0.2882 \\
        11 & JH             & 0.3678 & 0.2475 \\
        12 & Homa           & 0.3610 & 0.2993 \\
        13 & TSOTSALAB      & 0.1407 & 0.1012 \\
        \bottomrule
    \end{tabular}
    \caption{Average recall for both cases of qualitative evaluation.}
    \label{tab:recall_teams}
\end{table*}

\section{System Overview}

Our system is based on the two-stage approach to information retrieval. We consider the available texts of the documents (usually, just the title and the abstract) to be the \textit{queries}, while the subject definitions act as the \textit{documents} to be retrieved. In the first stage, we employ an approximate nearest neighbors (ANN) search algorithm~\cite{dasgupta2008random} over pre-computed subject embeddings. In the second stage, these $N$ retrieved subject descriptions are re-ranked using a cross-encoder model~\cite{Nogueira2019PassageRW}.

\subsection{First Stage}

The first stage of our system is a pre-trained bi-encoder model. We did not perform any additional fine-tuning, relying on the strength of
pre-trained multilingual embeddings, which have been shown to handle
cross-lingual retrieval effectively~\cite{dorkin-sirts-2024-sonajaht}. The only modification we made was to provide a task-specific prompt for the model. For each document, we queried the model with the document text and retrieved top $N$ subject descriptions. 

The model is only used once to obtain representations for each document and subject, which are then stored and reused as needed, making the model relatively cheap and fast to use. However, performing the semantic search over a large collection of subject representations remains computationally expensive. To mitigate that, we employed an approximate nearest neighbors algorithm to create a fast search index for the subjects as a data structure separate from the pre-computed representations. Accordingly, the index is then queried with document representations to retrieve subject candidates.

The first stage efficiently retrieves a coarse set of $N$ candidate subjects. The efficiency stems from leveraging ANN indices; while increasing the number of trees (n\_trees) enhances the index accuracy and thus retrieval quality (higher recall), it also increases memory usage and the time required to build and query the index. Varying this parameter provides flexibility in tuning Stage 1 performance against computational resources.

\subsection{Second Stage}

In the second stage, we employ a cross-encoder model to re-rank the subject candidates initially selected by the first-stage model. Specifically, pairs are formed where each instance consists of the target text and one candidate subject description. Each such pair is fed into the cross-encoder model.

This model functions as a classifier trained to assess the relevance or similarity between the paired texts (target text and candidate definition). It outputs a probability score indicating the likelihood of relevance, which we interpret directly as a normalized similarity score between 0 and 1. This scoring allows us to refine and reorder the list of subject candidates.

Cross-encoder models are well-suited for this task due to their proven ability to generalize across highly heterogeneous text collections~\citep{thakur2021beir,dorkin-sirts-2024-tartunlp-axolotl,10.1007/978-3-031-56063-7_10}. They excel at capturing subtle nuances by jointly processing document-query-like pairs to compute fine-grained relevance scores.

Given the specificity of this shared task problem, no suitable pre-trained off-the-shelf models existed. Consequently, we developed our cross-encoder model through a fine-tuning process: we took a large, multilingual text encoder pre-trained on general data and adapted it by training on the provided dataset.

While cross-encoders offer superior ranking performance—demonstrated by their ability to significantly boost recall (nearly doubling values in Table \ref{tab:recall_transposed}) compared to bi-encoder approaches—they are computationally expensive. This is primarily because generating predictions requires processing each candidate pair individually. This represents a key scalability consideration for this approach.

\section{Experimental setup}

The shared task data underwent minimal transformations to be suitable for our approach. For each document, the title and the abstract were concatenated. Similarly, the name and definition (when available) were concatenated for each subject. Thus, each document and each subject was represented by a single string, which served as input to our models.

Our solution primarily relied on the SentenceTransformers\footnote{\url{https://www.sbert.net/}} library for both stages of the system. We employed multilingual-e5-large-instruct\footnote{\url{https://huggingface.co/intfloat/multilingual-e5-large-instruct}}~\cite{wang2024multilingual} as the first stage model to produce document and subject representations. We customized the prompt used to encode the documents to include the following: \textit{``Instruct: Given the following title and abstract for the document, retrieve the relevant subjects classifying the document''}. The query prompt remained unchanged from the default prompt \textit{``Query:''}. We used the Annoy\footnote{\url{https://github.com/spotify/annoy}} library to build the approximate neighbors index with a somewhat large number of trees equal to \textbf{100} to maximize the recall at the cost of some performance. When selecting candidates for the second stage we the search\_k parameter to \textbf{50000} for the same purpose. Finally, for each document we selected $N$ equal to \textbf{512} subject candidates for re-ranking.

For the second stage, we fine-tuned a cross-encoder based on mdeberta-v3-base\footnote{\url{https://huggingface.co/microsoft/mdeberta-v3-base}}~\cite{he2021debertav3} with default parameters using SentenceTransformers. We trained on positive examples from document-subject pairs in the training set and constructed negative examples by pairing documents with randomly sampled subjects not explicitly linked to them. The model achieved high performance quickly during training; we stopped after one epoch as the F-score plateaued near 0.97. The resulting model was used to make predictions on the validation split. The model is available on HuggingFace\footnote{\url{https://huggingface.co/adorkin/llms4subjects-cross-encoder}}.

\section{Results}

\begin{table}[t]
\small
    \centering
    \begin{tabular}{lcc}
        \toprule
        \multirow{2}{*}{Recall at k} & \multirow{2}{*}{Bi-encoder} & Bi-encoder $\rightarrow$ \\
        & & Cross-encoder \\
        \midrule
        5  & 0.1161 & 0.2126 \\
        10 & 0.1555 & 0.2646 \\
        15 & 0.1773 & 0.2920 \\
        20 & 0.1932 & 0.3121 \\
        25 & 0.2080 & 0.3261 \\
        30 & 0.2399 & 0.3574 \\
        35 & 0.2496 & 0.3661 \\
        40 & 0.2590 & 0.3732 \\
        45 & 0.2655 & 0.3791 \\
        50 & 0.2719 & 0.3837 \\
        \bottomrule
    \end{tabular}
    \caption{Recall values per k for the submitted runs.}
    \label{tab:recall_transposed}
\end{table}

With our submission to the shared task, we aimed to build a hackathon-like proof-of-concept system to test the feasibility and measure the benefits of applying a two-stage information retrieval approach to match document contents with a structured knowledge resource. More specifically, our interest lied in the improvements attained by the second stage model at the cost of added complexity and computational requirements compared to using only bi-encoder representations for retrieval. 

With that purpose in mind, we submitted two runs for final evaluation: bi-encoder retrieval and two-stage retrieval. The results in Table \ref{tab:recall_transposed} demonstrate the substantial benefit of incorporating the cross-encoder re-ranking stage. Adding the second stage nearly doubles the recall across various cutoff points $k$, significantly improving performance over using only the Stage 1 bi-encoder retrieval. This highlights the effectiveness of leveraging sentence similarity between documents and subject definitions as relevance signals.

However, the enhanced performance comes with computational implications. Both stages involve parameters that directly impact scalability:

\begin{itemize}
    \item \textbf{Stage 1 (Bi-Encoder + ANN)}: While the initial candidate retrieval is relatively fast due to approximate nearest neighbor search, increasing the n\_trees parameter used in Annoy's index construction enhances recall by building more accurate indices. Additionally, the search\_k parameter determines how many nodes in the index are explored during the search, thus also improving recall. These improvements come at the cost of an increased memory footprint and longer indexing and querying times.
    \item \textbf{Stage 2 (Cross-Encoder)}: The primary computational bottleneck lies in applying the cross-encoder to re-rank all $N$ candidates for each document. This step is significantly more computationally demanding than Stage 1, especially as $N$ grows large. While feasible for annotating documents individually on local CPU hardware, deploying such a system at extreme scale—e.g., processing millions of documents in real-time or with strict resource limits—would necessitate careful management of parameters like n\_trees and $N$, potentially requiring optimization techniques (like quantization or using a shallow cross-encoder).
\end{itemize}

Upon examination of the results of the qualitative evaluation, we discover that the errors made by our system are primarily related to subjects with similar names and no definitions. The incorrectly assigned subject seems to be generally vaguely related to the document. However, they define a different subfield of a relevant subject.

For example, for an article titled \textit{``Model-based engineering of an automotive adaptive exterior lighting system: realistic example specifications of behavioral requirements and functional design''}, some of the tags that are considered correct are ``Automotive Engineering, Vehicle Construction, Conveyor Technology, Aerospace Technology`` and ``System Planning IT, Data Processing''. Meanwhile, ``Lighting technology Electrical engineering, Electrical power engineering: Calculation, Design and Construction of Lighting Systems'' and ``Outdoor Lighting Electrical Engineering, Electrical Power Engineering'' are only technically correct. Finally, ``Adaptive Process Model Measurement, Control and Regulation Technology'' and ``Model-Based Testing Computer Science, Data Processing'' are assigned incorrectly to this document by our system.

The main limitation of our approach is the reliance only on text representations. Using additional information such as related subjects and tag co-occurrence could have improved the performance of our system.

\section{Conclusion}

This paper described our solution to SemEval-2025 Task 5 based on a two-stage information retrieval system, where we used the documents to annotate as queries to retrieve candidate subject tags from a large collection of subject tags. For both stages we view sentence similarity between document texts and subject tag descriptions as the relevance score. Our system demonstrates moderate performance. However, we confirmed our hypothesis that a second-stage re-ranker substantially improves the performance of the system compared to using a bi-encoder as the only stage.

\section*{Acknowledgments}

This research was supported by the Estonian Research Council Grant PSG721.

\bibliography{custom}

\begin{thebibliography}{12}
\providecommand{\natexlab}[1]{#1}

\bibitem[{Asula et~al.(2021)Asula, Makke, Freienthal, Kuulmets, and Sirel}]{asula2021kratt}
Marit Asula, Jane Makke, Linda Freienthal, Hele-Andra Kuulmets, and Raul Sirel. 2021.
\newblock Kratt: developing an automatic subject indexing tool for the national library of estonia.
\newblock \emph{Cataloging \& Classification Quarterly}, 59(8):775--793.

\bibitem[{Dasgupta and Freund(2008)}]{dasgupta2008random}
Sanjoy Dasgupta and Yoav Freund. 2008.
\newblock Random projection trees and low dimensional manifolds.
\newblock In \emph{Proceedings of the fortieth annual ACM symposium on Theory of computing}, pages 537--546.

\bibitem[{Dorkin and Sirts(2024{\natexlab{a}})}]{dorkin-sirts-2024-sonajaht}
Aleksei Dorkin and Kairit Sirts. 2024{\natexlab{a}}.
\newblock \href {https://doi.org/10.18653/v1/2024.starsem-1.33} {S{\~o}najaht: Definition embeddings and semantic search for reverse dictionary creation}.
\newblock In \emph{Proceedings of the 13th Joint Conference on Lexical and Computational Semantics (*SEM 2024)}, pages 410--420, Mexico City, Mexico. Association for Computational Linguistics.

\bibitem[{Dorkin and Sirts(2024{\natexlab{b}})}]{dorkin-sirts-2024-tartunlp-axolotl}
Aleksei Dorkin and Kairit Sirts. 2024{\natexlab{b}}.
\newblock \href {https://doi.org/10.18653/v1/2024.lchange-1.11} {{T}artu{NLP} @ {AXOLOTL}-24: Leveraging classifier output for new sense detection in lexical semantics}.
\newblock In \emph{Proceedings of the 5th Workshop on Computational Approaches to Historical Language Change}, pages 120--125, Bangkok, Thailand. Association for Computational Linguistics.

\bibitem[{D'Souza et~al.(2025)D'Souza, Sadruddin, Israel, Begoin, and Slawig}]{dsouza-EtAl:2025:SemEval2025}
Jennifer D'Souza, Sameer Sadruddin, Holger Israel, Mathias Begoin, and Diana Slawig. 2025.
\newblock \href {https://aclanthology.org/2025.semeval2025-1.139} {Semeval-2025 task 5: Llms4subjects - llm-based automated subject tagging for a national technical library's open-access catalog}.
\newblock In \emph{Proceedings of the 19th International Workshop on Semantic Evaluation (SemEval-2025)}, pages 1082--1095, Vienna, Austria. Association for Computational Linguistics.

\bibitem[{He et~al.(2021)He, Gao, and Chen}]{he2021debertav3}
Pengcheng He, Jianfeng Gao, and Weizhu Chen. 2021.
\newblock \href {https://arxiv.org/abs/2111.09543} {Debertav3: Improving deberta using electra-style pre-training with gradient-disentangled embedding sharing}.
\newblock \emph{Preprint}, arXiv:2111.09543.

\bibitem[{Karpukhin et~al.(2020)Karpukhin, Oguz, Min, Lewis, Wu, Edunov, Chen, and Yih}]{karpukhin-etal-2020-dense}
Vladimir Karpukhin, Barlas Oguz, Sewon Min, Patrick Lewis, Ledell Wu, Sergey Edunov, Danqi Chen, and Wen-tau Yih. 2020.
\newblock \href {https://doi.org/10.18653/v1/2020.emnlp-main.550} {Dense passage retrieval for open-domain question answering}.
\newblock In \emph{Proceedings of the 2020 Conference on Empirical Methods in Natural Language Processing (EMNLP)}, pages 6769--6781, Online. Association for Computational Linguistics.

\bibitem[{Nogueira and Cho(2019)}]{Nogueira2019PassageRW}
Rodrigo Nogueira and Kyunghyun Cho. 2019.
\newblock Passage re-ranking with bert.
\newblock \emph{ArXiv}, abs/1901.04085.

\bibitem[{Petrov et~al.(2024)Petrov, MacAvaney, and Macdonald}]{10.1007/978-3-031-56063-7_10}
Aleksandr~V. Petrov, Sean MacAvaney, and Craig Macdonald. 2024.
\newblock Shallow cross-encoders for low-latency retrieval.
\newblock In \emph{Advances in Information Retrieval}, pages 151--166, Cham. Springer Nature Switzerland.

\bibitem[{Reimers and Gurevych(2019)}]{reimers-gurevych-2019-sentence}
Nils Reimers and Iryna Gurevych. 2019.
\newblock \href {https://doi.org/10.18653/v1/D19-1410} {Sentence-{BERT}: Sentence embeddings using {S}iamese {BERT}-networks}.
\newblock In \emph{Proceedings of the 2019 Conference on Empirical Methods in Natural Language Processing and the 9th International Joint Conference on Natural Language Processing (EMNLP-IJCNLP)}, pages 3982--3992, Hong Kong, China. Association for Computational Linguistics.

\bibitem[{Thakur et~al.(2021)Thakur, Reimers, R{\"u}ckl{\'e}, Srivastava, and Gurevych}]{thakur2021beir}
Nandan Thakur, Nils Reimers, Andreas R{\"u}ckl{\'e}, Abhishek Srivastava, and Iryna Gurevych. 2021.
\newblock Beir: A heterogenous benchmark for zero-shot evaluation of information retrieval models.
\newblock \emph{arXiv preprint arXiv:2104.08663}.

\bibitem[{Wang et~al.(2024)Wang, Yang, Huang, Yang, Majumder, and Wei}]{wang2024multilingual}
Liang Wang, Nan Yang, Xiaolong Huang, Linjun Yang, Rangan Majumder, and Furu Wei. 2024.
\newblock Multilingual e5 text embeddings: A technical report.
\newblock \emph{arXiv preprint arXiv:2402.05672}.

\end{thebibliography}

\end{document}